\documentclass[runningheads]{llncs}
\usepackage[T1]{fontenc}
\usepackage{graphicx}
\usepackage{amsmath} 
\usepackage{bm}
\usepackage{booktabs}
\usepackage{tabularx}
\usepackage{multirow}
\usepackage{marvosym}
\usepackage{xcolor}
\usepackage{xr-hyper}
\usepackage{hyperref}
\usepackage{orcidlink}
\begin{document}

\title{RecipeADQA: A User-aware Question-Answering Dataset for Multi-Constraint Recipe Recommendation
\thanks{Supported by KAKENHI (25K15130) Japan, and JST SPRING, Grant Number JPMJSP2132.}
}
\titlerunning{RecipeADQA}

\newcommand{\equalcontrib}{\textsuperscript{**}}
\author{Haotian Li \equalcontrib \orcidlink{0009-0001-7597-7444}
\and
Shuai Jiang \equalcontrib \orcidlink{0000-0002-3046-8689}
\and
Sayaka Kamei \orcidlink{0000-0003-1716-3028}
\and
Yasuhiko Morimoto \textsuperscript{(\Letter)} \orcidlink{0000-0001-7130-2864}
}

\authorrunning{H. Li et al.}

\institute{
Graduate School of Advanced Science and Engineering, Hiroshima University, Kagamiyama 1-7-1, Higashi-Hiroshima 739-8521, Japan\\
% \email{\{m252240, jiangshuai, s10kamei, morimo\}@hiroshima-u.ac.jp}
\email{morimo@hiroshima-u.ac.jp}
}

\maketitle

\begingroup
\renewcommand{\thefootnote}{\fnsymbol{footnote}}
\footnotetext[2]{These authors contributed equally to this work.}
\endgroup

\begin{abstract}

Question-answering recipe recommendation tasks lack datasets with ground-truth labels. Existing methods often generate datasets from simple combinations of ingredient or tag constraints, leading to unnatural expressions and unreasonable constraints that are hard to reflect users' real preferences.
This paper proposes a construction method for a benchmark-style QA dataset for natural-language recipe recommendation. Driven by real user behavior, our method integrates external user query and review data to extract fine-grained preference signals by LLMs and maps them onto a knowledge-graph-based tag schema, thereby constructing QA data consisting of preference and ingredient constraints. Besides, a greedy strategy for constraint selection is also used to control the answer size, while tag-based re-ranking is applied to improve data quality. The generated dataset RecipeADQA contains natural-language questions, structured constraints, and answer sets, resulting in 128,465 instances.
Experiments compare RecipeADQA with baseline datasets from two perspectives: data quality and retrieval-recommendation performance.
The results show that ours outperforms existing methods on the same recipe knowledge graph in coverage, higher question quality in naturalness and constraint rationality, and better queryability. Moreover, we design a noise-augmented retrieve-and-rerank model, in which a lightweight parser first retrieves candidate recipes, and a small language model then selects and ranks relevant recipes from a noisy candidate pool. The model trained on RecipeADQA outperforms the same model trained on the baseline datasets.
The relevant construction code and data have been open-sourced at a GitHub repository
~\footnote{\url{https://github.com/MorimoLab/RecipeADQA}}.

\end{abstract}

%-----------------------------------------------------------

\section{Introduction}

Recipe recommendation is a typical application scenario for conversational recommender systems.
Real users are likely to make compound natural language requests with their preference, such as ``a chicken dish for summer that is quick to make and can be cooked with an air fryer.'' Therefore, a question-answering dataset for recipe recommendation should cover a sufficient number of recipes and combinations of tags and ingredients, while ensuring that the questions closely reflect real user preferences and expression and that each question can be mapped to a queryable, verifiable answer set.

Existing recipe question-answering recommendation datasets are typically constructed using recipe knowledge graphs, yet significant limitations remain.
First, existing recipe knowledge graphs, such as FoodKG~\cite{haussmann2019foodkg}, lack user information, making it difficult to model real user preferences.
Second, although existing knowledge graphs contain a large number of tags, these tags are diverse in type, inconsistent in granularity, and lack systematic cleaning and structural organization.
As a result, existing dataset construction methods struggle to extract effective constraints that reflect real user preferences for question answering.
Meanwhile, existing synthetic construction methods can generate questions at scale. Still, they often rely on random tag combinations, leading to unnatural question expressions and constraint combinations that fail to support real user needs. For example, KGQA Benchmark~\cite{mohbat2025kerl} and PFoodReq~\cite{chen2021personalized} primarily generate questions based on nutritional tags or a small number of structured attributes. These datasets can verify whether a model recognizes explicit constraints, but they still have limitations in terms of question naturalness and the rationality of constraint combinations.

To address these issues, we propose a method to construct a natural-language recipe-recommendation QA Benchmark-style dataset, RecipeADQA.
The user-preference aspects and fine-grained directions are extracted from real user queries on ORCAS~\cite{craswell2020orcas} and from recipe reviews on Food.com~\cite{foodcom_kaggle_irkaal}, respectively, as prior knowledge for constructing question constraints.
Based on this, our enhanced recipe knowledge graph, RecipeADKG, is constructed by enriching the original tags in FoodKG and mapping them to our defined ``aspect::direction'' tag schema, thereby serving as constraints for generating structured QA pairs.
Finally, we use a greedy algorithm to combine constraints to keep the answer set at an appropriate size, and employ templates to convert structured QA pairs into natural-language question-and-answer data.
Experimental results demonstrate that RecipeADQA outperforms existing datasets in data quality and the utility of downstream retrieval recommendations. For details, RecipeADQA achieves higher RecipeADKG coverage than existing datasets, indicating it uses more KG data. Furthermore, we present a noise-augmented, retrieval-augmented generation recommendation task, and the model trained with RecipeADQA outperforms in distinguishing noise and producing effective recommendations.

The main contributions of this paper are as follows. 
\begin{enumerate}
    \item We design an aspect-direction labeling scheme grounded in real user queries and reviews. By treating external user behavior and review data as prior knowledge, the proposed scheme aligns existing recipe tags with preference aspects and their corresponding directions that reflect users’ actual concerns.
    \item We propose a systematic method for constructing the enhanced recipe knowledge graph, RecipeADKG, and the RecipeADQA question-answering dataset. The method enriches the recipe knowledge graph and leverages the aspect-direction labeling scheme to select and combine constraints for each question, yielding recipe question–answer pairs with rational constraint combinations, natural linguistic expressions, and controllable answer spaces.
    \item Experiments demonstrate that RecipeADQA achieves higher data quality than existing dataset construction methods and yields superior performance on downstream recipe retrieval and recommendation tasks.
\end{enumerate}

\section{Related Work}

\subsection{Recipe Datasets and Knowledge Graphs}

% Large-scale recipe corpora and food knowledge graphs provide essential data and knowledge foundations for research on food computing, recipe understanding, retrieval, and recommendation.
% Recipe1M+~\cite{salvador2017learning,marin2021recipe1m+} provides a large-scale multimodal corpus of recipe text, ingredients, instructions, and images, supporting cross-modal representation learning and retrieval. RecipeQA~\cite{yagcioglu2018recipeqa} formulates recipe understanding as a multimodal question-answering task, focusing on procedural comprehension and image--text alignment.
% These resources have advanced semantic representation learning for recipes, though they primarily focus on cross-modal retrieval and procedural understanding.

% Compared with recipe corpora composed mainly of unstructured text and images, food knowledge graphs explicitly organize entities and relations such as recipes, ingredients, nutrition, and dietary constraints.
% FoodKG~\cite{haussmann2019foodkg} links these elements with external ontologies, supporting structured retrieval, reasoning, and question answering. Recent approaches, such as RecipeRAG~\cite{loesch2025reciperag}, further integrate knowledge graphs with retrieval-augmented generation for personalized recipe recommendation.
% However, these resources generally lack QA instances grounded in natural-language recommendation needs, particularly unified alignment among questions, structured constraints, and answer sets. Consequently, generating trainable, verifiable QA data for recipe recommendation remains underexplored.

Large-scale recipe datasets and food knowledge graphs provide the foundation for understanding, retrieving, and recommending recipes.
Recipe1M+ \cite{salvador2017learning,marin2021recipe1m+} supports cross-modal learning over recipe text, ingredients, instructions, and images, while RecipeQA \cite{yagcioglu2018recipeqa} focuses on multimodal procedural understanding. FoodKG \cite{haussmann2019foodkg} further organizes recipes, ingredients, nutrition, and dietary information into a structured knowledge graph for retrieval and reasoning.
However, these resources are primarily recipe-centric, describing recipes and their attributes but offering only a limited representation of the multidimensional preferences users express in real-world recommendation scenarios.

\subsection{User-Aware Recipe Recommendation}

% Unlike general item recommendations, recipe recommendations involve multiple dimensions of user requirements, such as taste, ingredients, nutrition, and health restrictions. User-Aware modeling has therefore been an important research topic in food recommendation.
% Yum-me~\cite{yang2017yum} explored recommending foods that satisfy both user preferences and health objectives by jointly considering nutritional goals and dietary restrictions. Musto et al.~\cite{musto2020towards} proposed a holistic user model that integrates personal characteristics, dietary requirements, and food-domain knowledge to generate more personalized recipe recommendations.

% In addition to learning implicit preferences from historical interactions, several studies represent user requirements as explicit, verifiable constraints.
% Chen et al.~\cite{chen2021personalized} formulated personalized food recommendation as constrained question answering over a food knowledge graph, translating natural-language queries and dietary preferences into executable conditions. User-generated content also provides useful signals: Majumder et al.~\cite{majumder2019generating} leveraged user–recipe interactions and reviews to model personalized recipe generation.
% These works suggest that user preferences can be expressed as structured constraints, such as ingredients, nutrition, and dietary categories. However, limited work has systematically leveraged such preference evidence to scale up QA supervision, especially in realistic scenarios involving multiple heterogeneous conditions.

Recipe recommendations commonly involve multidimensional user requirements, including taste, ingredients, nutrition, health restrictions, and dietary preferences.
Yum-Me \cite{yang2017yum} learns fine-grained food preferences through user feedback and recommends meals that also satisfy nutritional goals, while Musto et al. \cite{musto2020towards} use holistic user profiles, including health characteristics and dietary constraints, to re-rank recipe recommendations. Majumder et al. \cite{majumder2019generating} model historical user--recipe interactions for personalized recipe generation. RecipeRAG \cite{loesch2025reciperag} also performs knowledge graph-based retrieval and generation under multiple user-specified constraints.
These studies demonstrate different ways of modeling and applying user preferences for developing recommendation or generation methods, but lack QA datasets whose constraint schemas are systematically derived from large-scale user query and review evidence.

\subsection{Recipe Retrieval and Recommendation QA Benchmarks}

In recent years, several studies have evaluated recipe retrieval and recommendation systems using natural-language questions or request descriptions. pFoodReQ~\cite{chen2021personalized} focuses on extracting recommendation-related conditions from user queries, while KERL~\cite{mohbat2025kerl} integrates knowledge-enhanced methods with large language models for tasks such as personalized recommendation and also constructs benchmark data with template-based questions and constraints.

However, these approaches primarily focus on request understanding, model evaluation, or template-based question generation. As a result, they rarely address three aspects simultaneously: (1) whether preference conditions are systematically grounded in large-scale user behavior and review evidence, (2) whether combinations of constraints reflect co-occurrence patterns in the knowledge graph, and (3) whether executable answer sets are explicitly controlled to remain suitable in size for training and evaluation.

% ---------------------------------------------

\section{Benchmark Dataset Construction}

We formulate recipe recommendation as a question-answering task with structured ingredient and tag constraints, and construct the benchmark dataset by integrating a recipe knowledge graph, real search queries, user reviews, and external recipe knowledge.
Compared with existing approaches, our construction method focuses on representing multidimensional user preferences extracted from external real search queries and user reviews. Then, such preferences are mapped into unified, structured ingredient and tag constraints to generate natural-language questions, thereby improving the rationality of constraint combinations and the naturalness of the questions.

\begin{figure}[htbp]
  \centering
  \includegraphics[width=\linewidth]{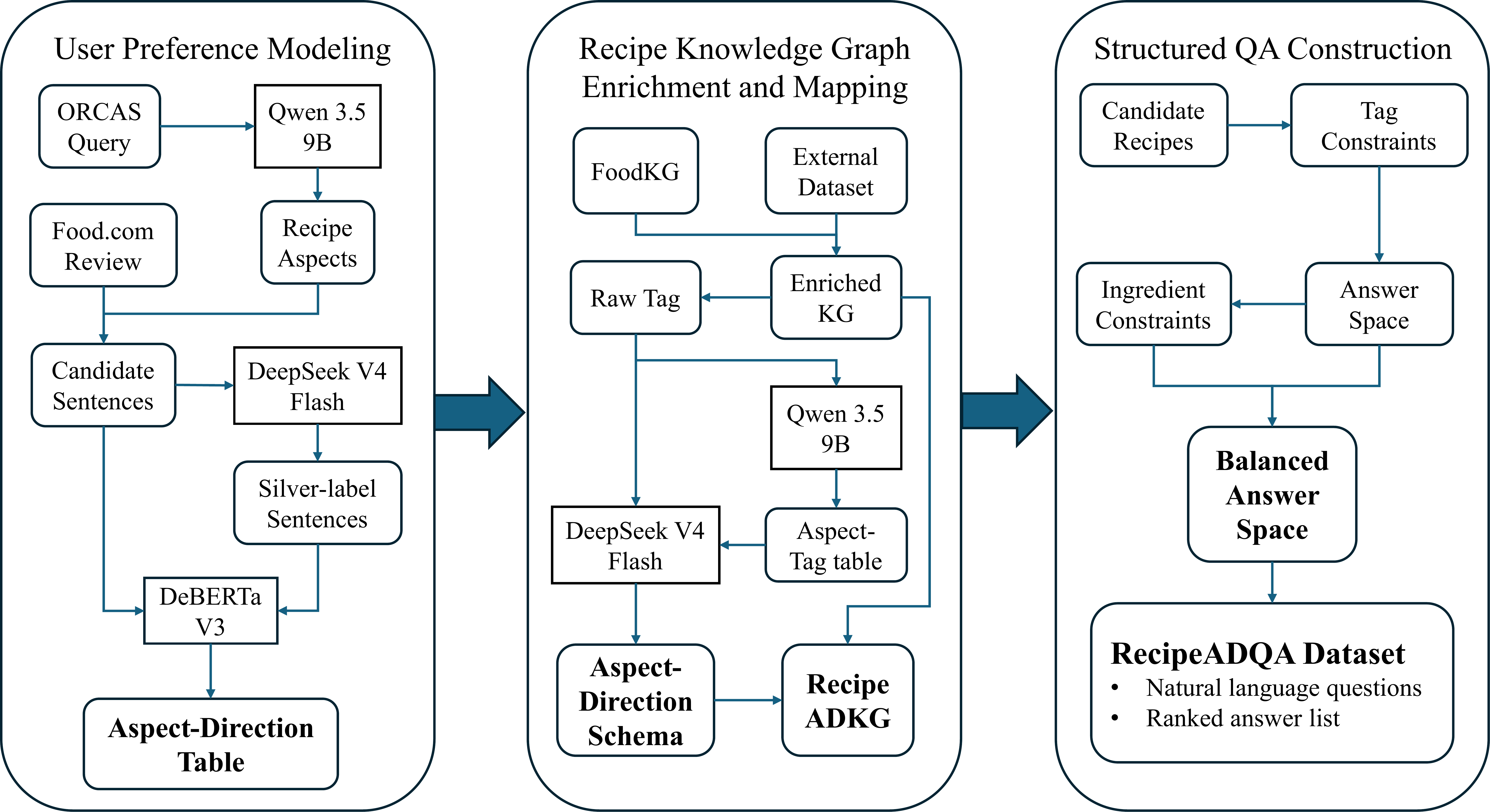}
  \caption{Overall workflow of the proposed method.}
  \label{fig:flow}
\end{figure}

Our benchmark dataset construction consists of three stages. First, we discover user-oriented aspects and preference directions from search queries and recipe reviews, respectively. Second, we map the original recipe tags in the recipe knowledge graph to a user-oriented ``aspect::direction'' tag schema, and enrich the graph to form the answer. Third, we generate structured QA pairs based on the tag schema and ingredients, and then convert the structured constraints into natural-language questions and answers for model training and evaluation. The overview of our construction method is shown in Figure~\ref{fig:flow}.

\subsection{Task Formulation}

Given a natural-language question $q$ and a candidate recipe collection $R$, the task is to retrieve recipes that satisfy the user's requirements. Each question $q$ is associated with an executable constraint set
$$
C(q) = \{c_1, c_2, \ldots, c_n\},
$$
where each $c_i$ represents an atomic requirement constraint, such as an aspect-direction tag or an ingredient inclusion or exclusion condition. The corresponding answer set is defined as
$$
A(q) = \{r \in R \mid \forall c_i \in C(q),\ r \models c_i\},
$$
where $r \models c_i$ indicates that recipe $r$ satisfies constraint $c_i$.

For example, the question ``I want a quick, low-sodium dinner suitable for children and free of nuts'' can be represented by constraints corresponding to \texttt{convenience::quick}, \texttt{health::low\_sodium}, \texttt{occasion::dinner}, \texttt{audience::kid\_friendly}, and the exclusion of nuts. Therefore, each instance in our dataset contains a natural-language question $q$, its structured constraint set $C(q)$, and a verifiable answer set $A(q)$. This alignment supports both natural-language recipe retrieval and the parsing of user questions into executable recommendation conditions.

\subsection{Query-driven and Review-based User Preference Modeling}

To identify the user's requirements and characterize the specific directions of preference for these requirements, we first introduce two kinds of data sources alongside the recipe dataset: user search queries and user reviews. This avoids relying solely on the subjective organization of raw tags, thereby aligning the constraints more closely with real user expressions.

\subsubsection{Query-driven Aspect Extraction}\label{asp_ext}

User search queries reflect the aspects of recipe preferences that users care about in real-world scenarios. Therefore, we first construct a recipe-related query set based on the ORCAS query-click dataset~\cite{craswell2020orcas}, comprising 18 million queries from real search engines, which provides real user preferences as prior knowledge. The original ORCAS dataset covers various types of webpages, with recipe pages constituting only a subset, and therefore cannot be used directly for recipe-specific analysis. Thus, we construct a whitelist of recipe-source websites based on known recipe domains from Recipe1M~\cite{salvador2017learning}, supplemented by manual curation and rule-based expansion. We finally deduplicate on the (qid, query) pair, resulting in a set of 67,886 recipe-related queries.
Based on this query set, we tokenize and normalize the query text to construct high-recall n-gram clues as supplementary assessments.
At the first stage of preference aspect extraction, we conducted multiple rounds of rolling tests on a small sample of the query set to calibrate the initial aspect lexicon, which was generated by GPT-5.5-Thinking through the ChatGPT web interface\footnote{
GPT-5.5-Thinking was accessed through ChatGPT on June 17, 2026.
},
to examine boundary issues and open labels, and refine the prompts and rules accordingly. Details are shown in Appendix~\ref{app-sec:aspe-disc}.
Next, we use Qwen3.5-9B~\cite{qwen3.5}, which the checkpoint obtained from Hugging Face\footnote{
\url{https://huggingface.co/Qwen/Qwen3.5-9B}
(accessed on April 8, 2026).
},
to perform semantic parsing on the query set and n-gram clues to obtain the aspect set:
$$
S = \{s_1, s_2, \dots, s_k\}
$$
where $k$ denotes the number of user requirement aspects identified from the query set, and each $s_i$ denotes a user requirement aspect, such as health, convenience, or occasion. The final retained aspect set $S$ contains the following $k=9$ categories that summarize common aspects in recipe queries:
%\begin{quote}
    \texttt{health, convenience, cooking method, taste, occasion, audience, quantity, region, tool}.
%\end{quote}

\subsubsection{Review-based Direction Inference}\label{dire_ext}

Although the ORCAS dataset provides user query signals, users often enter only a few keywords in web search, making queries insufficient to capture fine-grained preference directions for each aspect.
% Search queries help identify the aspects users care about, but they often do not fully specify the certain preference directions under each aspect.
Therefore, we further use the Food.com recipe reviews dataset~\cite{foodcom_kaggle_irkaal} to extract fine-grained direction preference under each aspect.
% Review texts contain rich natural evaluations of recipe attributes, such as whether a recipe is ``full of protein'', ``too sweet'', or ``needs more salt'' in the "health'' aspect. These expressions provide semantic evidence grounded in real user preferences for constructing the aspect-direction combination and mapping recipe tags.
We first build regular-expression patterns based on an aspect-direction lexicon generated by GPT-5.5; each aspect-direction pair is associated with a set of trigger words used to retrieve relevant candidate sentences from the full review corpus. For each review sentence $x$ triggered by the patterns, we define its corresponding candidate direction set as:
$$
D(x) = \left\{(s_i, d_j) \mid x \right\}
$$
where $d_j$ denotes $j$-th preference direction under $i$-th aspect $s_i$. For example, in the health aspect, directions may include low-sodium. We construct the aspect-direction-sentence set, totaling 4,418,496.
Next, we collect a stratified sample of 100,000 records based on direction and use DeepSeek-v4-flash\footnote{
API documentation: \url{https://api-docs.deepseek.com/}
(called between June 18 and June 23, 2026).
}~\cite{xu2026deepseek}
to annotate each sentence as ``support'', ``oppose'', or ``irrelevant'' for the Natural Language Inference (NLI) task, as the silver labels. The prompt template is shown in Appendix~\ref{app-sec:dire-anno}.
We formalize this annotation process as a classification function:
$$
f(x, s_i, d_j) \in \{\text{support}, \text{oppose}, \text{irrelevant}\}
$$
Based on these silver labels, we further split the data into training, validation, and test sets and fine-tune a DeBERTa-v3-base\footnote{
\url{https://huggingface.co/microsoft/deberta-v3-base}
 (accessed on May 26, 2026)
}~\cite{he2021debertav3} NLI classifier, ending with 0.811 test accuracy.
Since we restrict the scope of inference by the lexicon, we can infer over all sentences via fine-tuning a small language model at a lower cost than using LLMs.
After training, the model infers all aspect-direction-sentence sets. Finally, the system aggregates ``support'' and ``oppose'' evidence by aspect and direction, resulting in a direction evidence function:
$$
E(s_i, d_j) = N_{\text{support}}(s_i,d_j) - N_{\text{oppose}}(s_i,d_j)
$$
where $N_{\text{support}}(s_i,d_j)$ and $N_{\text{oppose}}(s_i,d_j)$ denote the number inferred as supporting or opposing the corresponding aspect-direction $(s_i,d_j)$, respectively. This value measures the strength of support for an aspect-direction combination in user reviews and is used to filter out low-support directions, as a reference for manual filtering. We finally obtained 393 aspect-direction pairs with stronger support, shown in Appendix~\ref{app-sec:adsc}.

\subsection{Recipe Knowledge Graph Enrichment and Mapping}

The recipe knowledge graph FoodKG~\cite{haussmann2019foodkg} is built on the extended Recipe1M dataset and includes recipes, ingredients, tags, and source information, for a total of 1,029,720 recipes. We use it for constructing question constraints and their answers. However, the 156,326 unique tags extracted from original recipe pages are loosely organized and contain noisy source, format, or metadata tags.
Besides, our aspect-direction pairs do not precisely match all tags.
Therefore, enrichment and normalized mapping are necessary.

We first enrich FoodKG to support more complex constraints for constraint mapping, including FoodData Central~\cite{fukagawa2022usda}, FNDDS~\cite{usda2020fndds}, and USDA ERS PP-NAP~\cite{carlson2020ppnap}, to establish links among ingredient weights, nutritional attributes, and price information.

At the same time, we convert the original tags in FoodKG into a unified ``aspect::direction'' tag schema.
First, we clean and deduplicate the original tag set $T_{raw}$ from FoodKG to obtain a normalized tag set $T_{clean}$ containing 10,101 tags. Then, we combine rule-based matching with coarse classification by Qwen3.5-9B to assign each tag in $T_{clean}$ to one of eleven classes: the 9 aspects extracted in Section~\ref{asp_ext}, ingredients, or other. The tag set $T$ containing matched aspects is forwarded to DeepSeek-v4-flash, which independently reclassifies them under stricter criteria.
After aspect classification, each tag is further mapped to the direction vocabulary obtained from Section~\ref{dire_ext} by DeepSeek-v4-flash.
For tags that cannot be mapped to any direction within their aspects, especially those in the ``other'' class, we perform cross-aspect mapping. Finally, we obtain a unified tag set:
$$
T_{final} = \{(s, d)\}
$$

We produced an aspect-direction tag schema containing 8,701 normalized FoodKG tags that map to 379 aspect-direction pairs, each of which contains at least one normalized FoodKG tag. Recipes that could not map to any aspect-direction pairs were filtered out, yielding a candidate recipe collection of 706,350.
Finally, each recipe is associated with at least one mapped aspect-direction tag, resulting 100\% recipe-tag coverage.
The specification comparison between RecipeADKG and FoodKG is shown in Table~\ref{tab:kg}.

\begin{table}[htbp]
\centering
\caption{Knowledge Graph Specification.}
\label{tab:kg}
\begin{tabular*}{0.65\linewidth}
  {@{\extracolsep{\fill}}lrr@{}}
    \toprule
    & \textbf{FoodKG} & \textbf{RecipeADKG} \\
    \midrule
    Total Recipes &  1,029,720 & 706,350 \\
    Unique Tags & 156,326 & 379  \\
    Recipe-Tag Coverage  & 90.31 & 100.00 \\
    \bottomrule
\end{tabular*}
\end{table}

\subsection{Structured QA Construction}

Based on the mapping, we can construct structured QA pairs.
First, we traverse all recipes $R$ in the knowledge graph as answer candidates, and select recipes with no more than 10 tags as seed recipes $R^{\prime}$ for question generation.
Unlike KERL~\cite{mohbat2025kerl}, which randomly samples tags, we prioritize using the complete tag set of each seed recipe as the question constraints, thereby reducing unnatural or unsatisfiable constraint combinations.
For each seed recipe $r_i$ in $R^{\prime}$, we extract the set of labels $T_{final}(r_i)$ under the final label scheme, and convert each label $t$ into a label constraint $c_t$. This yields a set of basic label constraints:
$$
C_{tag}(r_i) = \left\{c_t \mid t \in T_{final}(r_i) \right\}
$$
When the answer pool corresponding to a tag set is too large, the system adds ingredient inclusion or exclusion constraints $C_{ing}(r_i)$ to reduce the number of answers to a target range.
As shown in the Appendix~\ref{app-sec:cons-count}, we count the number of constraint combinations for all seed recipes. Since the average size of the answer space is kept at a suitable count of 25 when the number of constraints is 10, we determine the maximum number of constraints as 10, and the maximum number of answers as 25, so the size of the answer set for each question $q$ will satisfy $1 \leq |A(q)| \leq 25$.
Therefore, the set of structured constraints ultimately used to generate the questions is:
$$
C(r_i)=C_{tag}(r_i) \cup C_{ing}(r_i)
$$
After merging and filtering, we obtain 128,465 constraint combinations, which can be used to construct QA pairs. This algorithm is shown in Appendix~\ref{app-sec:answer}.

For several quantifiable directions under the \texttt{health} and \texttt{convenience} aspects, tag matching only determines whether a recipe satisfies tag requirements, but can not distinguish how strongly it satisfies. Therefore, we further re-rank the matched answers using the corresponding numeric attributes, such as nutritional values, cooking time, price, number of steps, and number of ingredients, as shown in Table~\ref{tab:rank_const}.

\begin{table}[]
    \centering
        \caption{Aspect-Direction Constraints for Re-ranking.}\label{tab:rank_const}
    \begin{tabularx}{\textwidth}{lX}
    \toprule
         Aspect ~&~ Direction \\
         \midrule
         \texttt{health} ~&~ high-protein, low-protein, low-sodium, low-fat, low-saturated-fat, low-cholesterol, low-carb, low-calorie, high-fiber, low-sugar, high-calcium\\
         \texttt{convenience} ~&~ quick (cooking minutes), affordable (price), easy (steps), few-ingredient \\
    \bottomrule
    \end{tabularx}
\end{table}

Finally, we convert the structured constraints into natural-language questions. We use different templates for each aspect to ensure consistency between questions and the structured constraints. Finally, we obtain 128,465 unique natural-language question-answering sets, each containing a natural-language question $q$, structured constraints $C(q)$, and a ranked set of answer recipes $A(q)$.

% ----------------------------------

\section{Experiments}

We evaluate our recipe question-answering dataset and baselines from two main perspectives:
\begin{itemize}
    \item Data Quality: To evaluate the dataset's coverage of the knowledge graph and further analyze the naturalness of the question texts and the rationality of constraint combinations, to verify whether the proposed construction method can generate high-quality QA pairs with reasonable coverage. This part is presented in Section~\ref{sec:quality}.
    \item Retrieval-Recommendation Performance: To evaluate the utility of different methods on the dataset, including conventional retrieval methods and the LLM-based retrieval-augmented generation (RAG) method, to assess the dataset's utility. This part is presented in Section~\ref{sec:retr} and \ref{sec:llm}.
\end{itemize}
To this end, we compare our dataset RecipeADQA with two baseline datasets. The first is a KGQA-style dataset that follows the QA generation strategy used in KERL~\cite{mohbat2025kerl}, and the second is a PFoodReq-style dataset~\cite{chen2021personalized}.
Since the original KERL implementation has not yet been fully open-sourced, we reproduce the KGQA-style dataset described in the paper for our mapped knowledge graph (i.e., RecipeADKG) and expand it to match our scale. Similarly, the PFoodReq-style dataset is reproduced in our knowledge graph using its open-source code; therefore, the scale is larger than that reported in its paper.
For fairness, we reproduce both datasets in our RecipeADKG.
Since RecipeADQA and the baselines are constructed from the same tag and ingredient information, RecipeADKG does not confer an exclusive advantage to our method through our cleaning and mapping; therefore, the performance differences mainly stem from their strategies for constraint selection and combination.
In the final setting, RecipeADQA and KGQA-style each contain 128,465 QA pairs, while PFoodReq-style contains 12,560 QA pairs, shown in Table~\ref{tab:cover}.

\begin{table}[ht]
\centering
\caption{Knowledge Graph Coverage Analysis.}
\label{tab:cover}
\begin{tabular*}{0.65\linewidth}
  {@{\extracolsep{\fill}}lrrrr@{}}
    \toprule
    \multirow{2}{*}{\textbf{Dataset}} &
    \multirow{2}{*}{\textbf{QA pairs}} &
    \multicolumn{3}{c}{\textbf{Coverage}} \\
    \cmidrule{3-5}
    & & Tag & Recipe & Edge \\
    \midrule
    PFoodReq-style & 12,560 & 43.54 & 10.28 & 13.58 \\
    KGQA-style & 128,465 & 3.69 & 29.42 & 35.37 \\
    RecipeADQA & 128,465 & 95.51 & 35.04 & 43.73 \\
    \bottomrule
\end{tabular*}
\end{table}

% ---
\subsection{Data Quality Evaluation}
\label{sec:quality}

\subsubsection{Knowledge Graph Coverage Analysis}

We first analyze the coverage of different datasets over RecipeADKG, including Tag Coverage, Recipe Coverage, and Edge Coverage. Specifically, Tag Coverage measures the number of mapped tags explicitly used as constraints in each QA dataset, reflecting the utilization of the tag space. Recipe Coverage is calculated by counting the number of distinct recipes across all answer sets in each dataset and computing their proportion relative to all recipes in RecipeADKG, reflecting the extent to which the dataset covers recipe entities. Edge measures the coverage of tags, ingredient nodes, and their associated edges adjacent to the answer recipes in the knowledge graph, thus characterizing the scale of the subgraph covered by the answer sets.

As shown in Table~\ref{tab:cover}, RecipeADQA achieves the highest coverage across all three dimensions, including 95.51\% tag coverage, indicating that the proposed construction strategy uses a substantially broader portion of the normalized knowledge graph.
KGQA-style covers many recipes but only a few tags, because its construction scope is limited to a small set of health-related tags. In contrast, PFoodReq-style has limited recipe and edge coverage due to its smaller, more restricted constraint space.

% ---
\subsubsection{Question Quality Evaluation}\label{sec:quest}

Through the QA itself, we explore whether high coverage can lead to higher-quality question construction.
To evaluate question quality, we randomly sample questions from each dataset and design a questionnaire template (shown in Appendix~\ref{app-sec:questionnaire}) for respondents to rate each generated question on a scale of 1 to 5 along two dimensions: Naturalness and Rationality.
First, we conduct an automatic, blind evaluation using the LLM DeepSeek-V4-Flash, which rates 1,000 generated questions for each dataset.
To further examine human perceptions, we invited 20 respondents aged 24 to 28 who held bachelor's degrees to conduct a manual evaluation.
Each participant rated 9 recipe questions, with 3 questions sampled from each dataset, for a total of 180 questions. Same with LLM, for each question, the participant assigned separate scores for naturalness and plausibility, both on the same 1-5 scale.

\begin{table}[htbp]
\centering
\caption{Questionnaire Evaluation Results.}
\label{tab:quest}
\begin{tabular*}{\linewidth}
  {@{\extracolsep{\fill}}lcccccccc@{}}
    \toprule
    \multirow{3}{*}{Dataset} 
    & \multicolumn{4}{c}{LLM} 
    & \multicolumn{4}{c}{Human} \\
    \cmidrule(lr){2-5} \cmidrule(lr){6-9}
    & \multicolumn{2}{c}{Naturalness}
    & \multicolumn{2}{c}{Rationality}
    & \multicolumn{2}{c}{Naturalness}
    & \multicolumn{2}{c}{Rationality} \\
    \cmidrule(lr){2-3} \cmidrule(lr){4-5}
    \cmidrule(lr){6-7} \cmidrule(lr){8-9}
    & mean & median & mean & median & mean & median & mean & median \\
    \midrule
    PFoodReq-style & 2.166 & 2 & 2.560 & 3 & 2.467 & 2 & 2.533 & 3 \\
    KGQA-style & 2.193 & 2 & 2.239 & 2 & 2.817 & 3 & 2.583 & 3 \\
    RecipeADQA
    & \textbf{3.642} & \textbf{4} & \textbf{4.069} & \textbf{4} 
    & \textbf{3.500} & \textbf{4} & \textbf{3.433} & \textbf{4} \\
    \bottomrule
\end{tabular*}
\end{table}

The results in Table~\ref{tab:quest} show that RecipeADQA consistently receives the highest naturalness and rationality scores in both LLM-based and human evaluations. The agreement between the two evaluation settings indicates that its improved constraint coverage does not come at the cost of unnatural expressions or implausible constraint combinations.
The case study in the Appendix~\ref{app-sec:qa-examp} also demonstrates this.

% ---
\subsection{Retrieval Recommendation Evaluation}
\label{sec:retr}

We use BM25 and BGE-small-en-v1.5~\cite{xiao2024c} as retrieval-only baselines to examine whether the recipe recommendation task can be solved by simple lexical matching or semantic vector retrieval, which serve as lower bounds for the natural language recommendation task.
For a fair comparison, we conduct this experiment using the same 1,000 generated data for each dataset as in Section~\ref{sec:quest}.
BM25 is a classic sparse retrieval method based on word frequency and inverse document frequency, whereas BGE is a dense vector retrieval model based on a pre-trained language model.
Because answer set sizes vary substantially across datasets, we report only two metrics: Hit Rate and Mean Reciprocal Rank (MRR). Hit Rate measures whether at least one correct answer appears in the returned Top-K results, while MRR measures the ranking position of the first correct answer. For both of these, the higher, the better.

\begin{table}[htbp]
\centering
\caption{Retrieval Results.}
\label{tab:retri}
\begin{tabular*}{0.65\linewidth}
  {@{\extracolsep{\fill}}lcccc@{}}
    \toprule
    \multirow{2}{*}{\textbf{Dataset}} 
    & \multicolumn{2}{c}{\textbf{BM25}} 
    & \multicolumn{2}{c}{\textbf{BGE}} \\
    \cmidrule(lr){2-3} \cmidrule(lr){4-5}
    & Hit Rate & MRR & Hit Rate & MRR \\
    \midrule
    PFoodReq-style & 37.7 & 17.40 & 23.4 & 9.71 \\
    KGQA-style & 26.7 & 16.30 & 21.9 & 10.99 \\
    RecipeADQA & \textbf{54.3} & \textbf{33.22} & \textbf{28.7} & \textbf{14.70} \\
    \bottomrule
\end{tabular*}
\end{table}

The results in Table~\ref{tab:retri} show that RecipeADQA achieves the strongest performance with both BM25 and BGE, suggesting that its questions preserve explicit lexical and semantic cues consistently associated with the corresponding answer sets. Nevertheless, the overall scores remain modest, particularly for dense retrieval, indicating that generic similarity matching is insufficient to enforce multiple constraints simultaneously. These results motivate the structured parsing and candidate re-ranking evaluation in the following section.

% -------------------------------------------
\subsection{LLM Recommendation Evaluation}
\label{sec:llm}

\begin{figure}[htbp]
  \centering
  \includegraphics[width=0.85\linewidth]{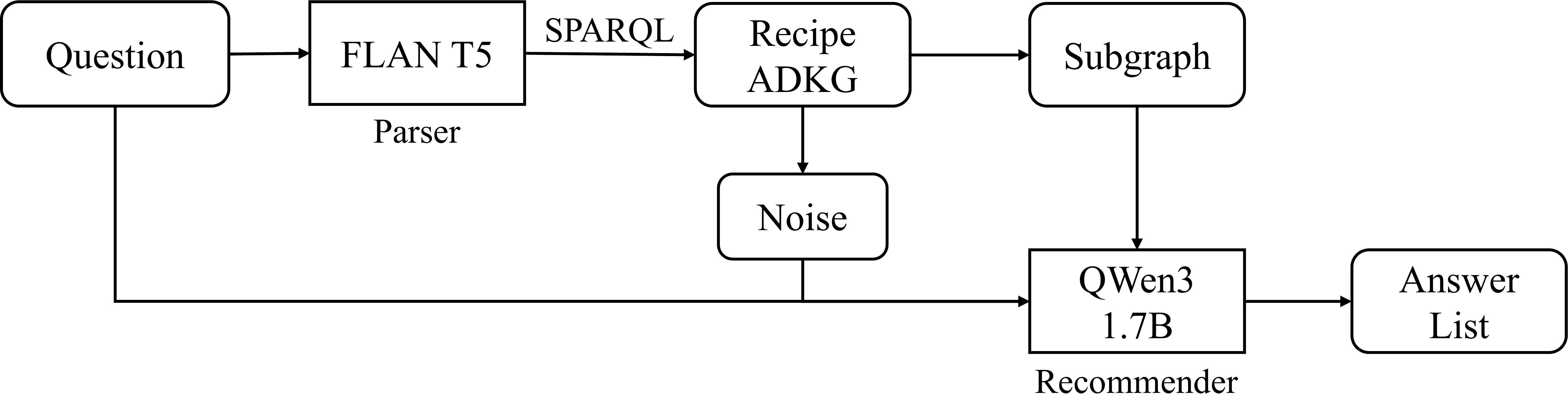}
  \caption{Noise-augmented LLM-based RAG recommendation framework.}
  \label{fig:rag}
\end{figure}

In real-world retrieval-augmented recommendation scenarios, the initial retrieval results often include candidate recipes that do not fully satisfy all user requirements. To simulate this situation, we here design a noise-augmented LLM-based RAG recommendation framework that adds a specified number of noisy recipes to the ground-truth answer set. The model should identify and rank recipes from the mixed candidate set that meet the user’s needs, based on multiple constraints expressed in the query.
To investigate learning performance across different datasets, we randomly sample 10,000 QA pairs from each dataset and split them into training, validation, and test sets in an 8:1:1 ratio. The overall structure is shown in Figure~\ref{fig:rag}.

At the natural language retrieval stage, we fine-tune the FLAN-T5~\cite{chung2024scaling} model to parse natural-language questions into executable, structured SPARQL queries. The queries are then executed on the knowledge graph to retrieve answer sets that satisfy all constraints. The training results are that the model achieves over 99\% parsing accuracy on all datasets. This indicates that the generated questions are well-parsable and can be reliably mapped to structured queries.

Next, we feed the natural-language question, the list of retrieved answer recipes, and several noise recipes into the backbone LLM recommender to select and rank recommendations from the candidate set.
To ensure fairness and control the scale of the experiment, we apply a uniform truncation to the list of answers. Specifically, for each sample, we first collect all correct answers that satisfy the constraints and truncate them to a maximum of 25.
Then, we augment the candidate set to 30 items with noise-recipe samples, prioritizing recipes that violate at most one constraint to ensure high similarity to the correct answers.
Based on this candidate set, the model must output up to 25 ordered results that satisfy the constraints specified in the question. We use Qwen3-1.7B~\cite{yang2025qwen3}, with the checkpoint obtained from Hugging Face\footnote{
\url{https://huggingface.co/Qwen/Qwen3-1.7B}
(accessed on June 23, 2026).
},
as the backbone recommender with LoRA, and each experiment is trained on a single data source. The prompt template for training is shown in Appendix~\ref{app-sec:rag-prom}.
% We evaluate performance at various Top-K using F1, NDCG, and Noise.
% Specifically, Noise represents the percentage of test instances whose Top-$K$ recommendation list contains at least one noise recipe. A lower value is better, indicating that the model is less likely to include noise recipes in its recommendations.
% For details, TP here refers to the number of intersections between the model's top-K outputs and the top-K ground-truth answers, which is used to calculate the F1 score.

We trained the same recommendation model on different QA datasets and evaluated performance using Precision, Recall, F1, and Noise, shown in Table~\ref{tab:rag}.
Precision measures the average proportion of true recipe answers among those recommended for each question; Recall measures the average coverage of true recipe answers for each question; and F1 provides a combined measure of both. All three metrics were calculated per question and then macro-averaged.
Noise represents the proportion of questions for which the model's output included at least one noise recipe; a lower value indicates better noise recognition performance.
For baseline datasets, we also reproduced them using the original FoodKG (denoted by -FoodKG suffix) and selected an equal amount of data for model training to investigate how the knowledge graph influences the recommendation model's learning process.
Furthermore, since only RecipeADQA provides recipe rankings, we truncated the recommendation lists generated by a single model inference at K=5, 10, and 25, respectively, and calculated NDCG@K; the results are presented in Table~\ref{tab:rag_ndcg}.
Besides, to account for variations in the length of the real recipe answer space across datasets, we used the Macro F1 score as the primary metric and conducted stratified experiments on each dataset based on the number of answers; details are provided in the Appendix~\ref{app-sec:rag-result}.

\begin{table}[ht]
\centering
\caption{Recommendation Results.}
\label{tab:rag}
\begin{tabular*}{0.8\linewidth}
  {@{\extracolsep{\fill}}lcccc}
    \toprule
    \textbf{Dataset} & \textbf{Precision} & 
    \textbf{Recall} & \textbf{F1} & 
    \textbf{Noise$\downarrow$} \\
    \midrule
    PFoodReq-FoodKG
    & 91.02 & 91.00 & 88.39 & 13.0 \\
    PFoodReq-RecipeADKG
    & 87.15 & 92.78 & 87.14 & 27.5 \\
    KGQA-FoodKG
    & 98.21 & \textbf{99.17} & 98.42 & 10.4 \\
    KGQA-RecipeADKG
    & 88.92 & 93.18 & 89.21 & 41.3 \\
    \midrule
    RecipeADQA
    & \textbf{99.36} & 98.79 & \textbf{99.04} & \textbf{8.9} \\
    \bottomrule
\end{tabular*}
\end{table}

\begin{table}[ht]
\centering
\caption{Ranking Results for RecipeADQA.}
\label{tab:rag_ndcg}
\begin{tabular*}{0.7\linewidth}
  {@{\extracolsep{\fill}}lccc}
    \toprule
    \textbf{Dataset}
    & \textbf{NDCG@5}
    & \textbf{NDCG@10}
    & \textbf{NDCG@25}\\
    \midrule
    RecipeADQA
    & 95.25 & 97.39 & 98.61\\
    \bottomrule
\end{tabular*}
\end{table}

The results show that the model trained on RecipeADQA achieved the highest Precision and F1 scores and the lowest Noise, indicating its ability to filter noisy recipes while preserving a high proportion of correct recommendations.
Although KGQA on the original FoodKG achieved the highest Recall, its lower Precision and F1 scores, along with higher Noise, suggest that this Recall gain came at the cost of more incorrect selections.
%After reconstruction on RecipeADKG, both KGQA and PFoodReq showed improved Precision and F1 scores and substantially reduced Noise, reflecting a trade-off between stricter noise control and slightly lower answer coverage.
In addition, Appendix~\ref{app-sec:rag-result} shows that RecipeADQA consistently outperforms the other datasets in stratified experiments.
Its NDCG scores at Top-5, Top-10, and Top-25 were 95.25\%, 97.39\%, and 98.61\%, respectively, further demonstrating stable ranking quality across different evaluation depths.

\section{Conclusion}
In this paper, we presented RecipeADQA, a benchmark-style QA dataset for natural-language recipe recommendation, together with the enhanced knowledge graph RecipeADKG. 
By extracting preference aspects and directions from real user queries and reviews, normalizing and mapping raw recipe tags with varying semantics and granularities into a unified aspect-direction schema, and controlling answer spaces through tag and ingredient constraints, our method constructs 128,465 QA instances with natural-language questions, executable constraints, and verifiable ranked answers.
Experiments show that RecipeADQA achieves higher knowledge graph coverage, better question naturalness and constraint rationality, and stronger retrieval and recommendation performance than the KGQA-style and PFoodReq-style baselines. It also enables models to more effectively distinguish relevant recipes from highly similar noise candidates.
Future work will explore how RecipeADQA can support the development of stronger constraint-aware recipe recommendation systems, including more effective retrieval, ranking, reasoning, and personalized recommendation methods.

\paragraph{AI-Assisted Writing Disclosure.}
GPT-5.5-Thinking, accessed through the ChatGPT web interface between June 24 and July 18, 2026, was used to assist with the language polishing of the manuscript and supplementary appendix. The authors reviewed and revised all AI-assisted text and take full responsibility for the accuracy, integrity, and originality of the submitted material.

\bibliographystyle{splncs04}
\bibliography{ref}

\end{document}

\title{Appendix}
\author{}
\institute{}
\maketitle

\appendix

\section{Aspect Discovery}
\label{sec:aspe-disc}

Table~\ref{tab:aspe-disc} lists the aspect families extracted in the rolling aspect discovery, summary at the 50 queries, 1,000 queries, 5,000 queries, and full queries stages. Names are retained exactly as recorded at each stage. 

\begin{table}[]
    \centering
    \small
    \caption{Aspect-family labels extracted in rolling aspect discovery outputs at each query scale.}
    \label{tab:aspe-disc}
    \begin{tabularx}{\textwidth}{cXc}
        \toprule
        Stage & Aspect-family labels & Count \\
        \midrule
        50 queries & convenience; cuisine\_region; ingredient\_inclusion; method; occasion\_or\_meal; quality\_subjective; serving\_quantity; tool & 8 \\
        1,000 queries & audience; convenience; cuisine\_region; diet\_health; ingredient\_exclusion; ingredient\_inclusion; method; occasion\_or\_meal; quality\_subjective; source\_brand; taste\_texture;tool & 12 \\
        5,000 queries & audience; convenience; cuisine\_region; diet\_health; ingredient\_exclusion; ingredient\_inclusion; method; occasion\_or\_meal; quality\_subjective; serving\_quantity; source\_brand; taste\_texture; time\_temperature; tool & 14 \\
        Full queries & audience; convenience; food\_pairing; health; ingredient\_brand; ingredient\_exclusion; ingredient\_inclusion; method; occasion; quality\_subjective; quantity; region; taste; tool & 14 \\
        \bottomrule
    \end{tabularx}
\end{table}

These labels are intermediate parser families. We normalized labels such as \texttt{cuisine\_region} to \texttt{region}, \texttt{diet\_health} to \texttt{health}, and \texttt{taste\_texture} to \texttt{taste}. Labels representing ingredient, brands, generic quality, time or temperature, food pairing, or unresolved other cases were not retained as standalone aspects. This process produced the nine final aspects used by RecipeADQA: 
\begin{quote}
    health, convenience, method, taste, occasion, audience, quantity, region, and tool.
\end{quote}

% ----------------------------------------------------------------

\section{Direction Annotation}
\label{sec:dire-anno}

% \subsection{Silver-Label Annotation}
% \label{ssec:silver}

We used DeepSeek-v4-flash to annotate the stratified review sample with the silver labels \texttt{support}, \texttt{oppose}, and \texttt{irrelevant}. The annotation prompt is shown in Prompt~\ref{pmpt:dir-silv}.

\Needspace{24\baselineskip}
\begin{promptfloat}[htbp]
\begin{lstlisting}
[SYSTEM]
Judge whether one recipe review sentence supports the target preference.
Return JSON only: {"label":"support|oppose|irrelevant","confidence":0-1,"rationale":"short"}.

Input:
- target_preference: what the reviewer prefers.
- sentence: the evidence sentence. Judge only this sentence.

Labels:
- support: the sentence supports the target preference.
- oppose: the sentence contradicts that preference or supports the opposite.
- irrelevant: the sentence does not provide evidence about the target preference.

Distinguish factual statements from preference evidence: a sentence that only states something was used, made, served, cooked, or mentioned is not enough by itself. Do not infer from recipe title or outside context.

[USER]
{
  "sample_id": <SAMPLE_ID>,
  "aspect": "<ASPECT>",
  "direction": "<DIRECTION>",
  "target_preference": "<TARGET_PREFERENCE>",
  "sentence": "<EVIDENCE_SENTENCE>"
}
\end{lstlisting}
\caption{DeepSeek-v4-flash NLI silver-label annotation prompt.}
\label{pmpt:dir-silv}
\end{promptfloat}

% ----------------------------------------------------------

\section{Aspect-Direction Pairs}
\label{sec:adsc}

The Table~\ref{tab:aspect-direction-inventory} contain all 393 Aspect-Direction pairs.

{
\small
\begin{xltabular}{\textwidth}{
    @{}
    l
    >{\raggedright\arraybackslash}X
    @{}
}
\caption{Complete aspect--direction inventory}
\label{tab:aspect-direction-inventory}\\

\toprule
Aspect & Directions \\
\midrule
\endfirsthead

\caption[]{Complete aspect--direction inventory (continued)}\\
\toprule
Aspect & Directions \\
\midrule
\endhead

\midrule
\multicolumn{2}{r@{}}{\small Continued on next page}\\
\endfoot

\bottomrule
\endlastfoot

audience &
adults; athletes; babies; beginner cooks; bodybuilders; busy people; children; expecting mothers; families; fat-loss dieters; fitness enthusiasts; guests; gym-goers; inexperienced cooks; meal preppers; office workers; parents; picky eaters; post-workout eaters; pre-workout eaters; pregnant women; protein-focused eaters; seniors; seniors with chewing difficulty; shift workers; students; teenagers; toddlers; travelers
\\

\addlinespace
convenience &
accessible; affordable; easy; equipment-light; few-ingredient; hands-off; low-cleanup; low-prep; make-ahead; meal-prep-friendly; no-cook; quick; storage-friendly
\\

\addlinespace
health &
allergen-free; anti-inflammatory; balanced; blood-pressure-friendly; clean; dairy-free; diabetic-friendly; easy-to-digest; egg-free; gluten-free; gut-friendly; halal; healthy; heart-healthy; high-calcium; high-fiber; high-protein; immune-supporting; keto; kidney-friendly; kosher; light; low-FODMAP; low-calorie; low-carb; low-cholesterol; low-fat; low-purine; low-saturated-fat; low-sodium; low-sugar; non-alcoholic; nut-free; nutrient-dense; plant-forward; reflux-friendly; soy-free; vegan; vegetarian; weight-gain-friendly; weight-loss-friendly; whole-food-based; whole-grain
\\

\addlinespace
method &
air-fried; baked; blackened; blanched; boiled; braised; broiled; caramelized; charred; cured; deep-fried; fried; grilled; marinated; microwaved; pan-fried; poached; pressure-cooked; raw; roasted; sautéed; seared; simmered; slow-cooked; smoked; sous-vided; steamed; stewed; stir-fried; stovetop; toasted
\\

\addlinespace
quantity &
family-size; five-person; four-person; large-serving; one-person; scaled-batch; small-serving; three-person; two-person
\\

\addlinespace
occasion &
anniversary; appetizer; baby-shower; barbecue; birthday; bread; breakfast; brunch; camping; christmas; date-night; dessert; dinner; dinner-party; diwali; drink; easter; eid; fall; family-gathering; fathers-day; game-day; graduation; halloween; hanukkah; holiday; independence-day; leftover-meal; lunar-new-year; lunch; main-course; mothers-day; new-year; one-dish-meal; party; passover; picnic; potluck; ramadan; salad; sauce; school-day; side-dish; snack; soup; spring; st-patricks-day; summer; thanksgiving; valentine; wedding; weekday; weekend; winter; workday
\\

\addlinespace
region &
Algerian-style; American-style; Argentinian-style; Asian-style; Australian-style; Austrian-style; Brazilian-style; British-style; Cajun-style; Canadian-style; Cantonese-style; Caribbean-style; Chilean-style; Chinese-style; Colombian-style; Creole-style; Cuban-style; Czech-style; Danish-style; Egyptian-style; English-style; Ethiopian-style; Filipino-style; Finnish-style; French-style; German-style; Greek-style; Hawaiian-style; Hungarian-style; Icelandic-style; Indian-style; Indonesian-style; Iranian-style; Irish-style; Israeli-style; Italian-style; Jamaican-style; Japanese-style; Jewish-style; Korean-style; Lebanese-style; Malaysian-style; Mediterranean-style; Mexican-style; Middle Eastern-style; Moroccan-style; New Zealand-style; Nigerian-style; Nordic-style; Norwegian-style; Ontario-style; Pakistani-style; Palestinian-style; Persian-style; Peruvian-style; Polish-style; Portuguese-style; Puerto Rican-style; Punjabi-style; Quebec-style; Roman-style; Russian-style; Scandinavian-style; Scottish-style; Sichuan-style; Sicilian-style; South African-style; South Indian-style; Southern-style; Spanish-style; Swedish-style; Syrian-style; Tex-Mex-style; Thai-style; Tunisian-style; Turkish-style; Tuscan-style; Ukrainian-style; Vietnamese-style
\\

\addlinespace
taste &
aromatic; bitter; bland; brothy; buttery; caramelized; cheesy; chewy; chunky; comforting; creamy; creamy-and-crunchy; crispy; crispy-outside-soft-inside; crumbly; crunchy; crusty; custardy; dense; dry; earthy; firm; flaky; flavorful; fluffy; fresh; fruity; garlicky; gooey; greasy; herby; juicy; less-greasy; less-salty; less-sweet; light; meaty; mild; moist; mushy; nutty; rich; roasted; salty; salty-and-sweet; saucy; savory; smoky; smooth; smooth-and-creamy; soft; sour; spicy; sticky; sweet; sweet-and-sour; sweet-and-spicy; tangy; tender; thick; thin; tough; umami-rich
\\

\addlinespace
tool &
Dutch oven; air fryer; baking dish; baking tray; blender; bread machine; cake pan; candy thermometer; casserole dish; convection oven; crockpot; cutting board; dehydrator; foil; food processor; garlic press; grater; griddle; grill; hand mixer; hot plate; ice cream maker; kitchen scale; knife; ladle; lid; loaf pan; mandoline; microwave; mixing bowl; mortar and pestle; muffin tin; oven; panini press; parchment paper; pasta machine; peeler; pie dish; piping bag; pizza stone; pressure cooker; ramekin; refrigerator; rice cooker; roasting pan; rolling pin; salad bowl; sandwich maker; saucepan; scone cutter; scone pan; sifter; skillet; slow cooker; smoker; spatula; stand mixer; steamer; stockpot; stovetop; strainer; thermometer; toaster oven; tongs; toothpick; waffle maker; whirley pop; whisk; wire rack; wok; zester
\\

\end{xltabular}
}

% -------------------------------------------------------------

\section{Constraint Combination Analyze}
\label{sec:cons-count}

% 我们分析了RecipeADKG中全部菜谱的tag约束组合数量，以及符合的答案菜谱空间，如表1所示。可以看到，当约束组合数达到10时，平均答案数量达到一个相对合理的数值为25。因此我们将约束上限设置为10，最大答案数量限制为5，按章节1的算法进行答案空间控制。最终的约束和答案数量如表2所示。
We analyzed the number of tag-constraint combinations across all recipes in RecipeADKG and the corresponding answer recipe space, as shown in Table~\ref{tab:raw-cons}. It can be seen that when the number of constraint combinations reaches 10, the average number of answers reaches a relatively reasonable value of 25. Therefore, we set the upper limit of the constraint to 10, limit the maximum number of answers to 25, and control the answer space according to the Algorithm~\ref{lst:answer_control}. The final count of constraints and answers is shown in Table~\ref{tab:cont-cons}.

\begin{table}[htbp]
\centering
\small
\caption{Initial constraint combination counts and answer space sizes.}
\label{tab:raw-cons}
\begin{tabular}{rrrrr}
\toprule
Constraint & Constraint count & Max answer count & Avg. answer count & Min answer count \\
\midrule
1 & 273 & 363,800 & 13,348.78 & 1 \\
2 & 2,644 & 213,389 & 2,961.42 & 1 \\
3 & 6,437 & 70,673 & 1,296.16 & 1 \\
4 & 11,646 & 40,748 & 626.12 & 1 \\
5 & 19,843 & 22,850 & 323.12 & 1 \\
6 & 29,410 & 16,574 & 172.26 & 1 \\
7 & 34,858 & 12,641 & 96.15 & 1 \\
8 & 35,433 & 9,862 & 57.02 & 1 \\
9 & 31,620 & 6,110 & 36.69 & 1 \\
10 & 26,910 & 5,088 & 24.98 & 1 \\
\bottomrule
\end{tabular}
\end{table}

\begin{table}[htbp]
\centering
\small
\caption{Constraint combination counts and answer space sizes after answer control.}
\label{tab:cont-cons}
\begin{tabular}{rrrrr}
\toprule
Constraint & Constraint count & Max answer count & Avg. answer count & Min answer count \\
\midrule
1 & 38 & 25 & 8.342 & 1 \\
2 & 618 & 25 & 10.539 & 1 \\
3 & 2,981 & 25 & 13.647 & 1 \\
4 & 7,251 & 25 & 15.340 & 1 \\
5 & 15,860 & 25 & 15.527 & 1 \\
6 & 25,906 & 25 & 14.219 & 1 \\
7 & 35,998 & 25 & 12.946 & 1 \\
8 & 39,404 & 25 & 11.155 & 1 \\
9 & 35,643 & 25 & 8.988 & 1 \\
10 & 29,180 & 25 & 6.853 & 1 \\
\bottomrule
\end{tabular}
\end{table}

% -------------------------------------------------------------

\section{Algorithm for Answer Control}
\label{sec:answer}

\Needspace{32\baselineskip}
\begin{lstlisting}[language=Python, caption={Constraint filtering and answer space control.}, label={lst:answer_control}]
Input:
    R           # candidate recipes in the knowledge graph
    B = 10      # maximum number of total constraints
    U = 25      # maximum accepted answer size
Output:
    Q           # structured constraints

for each recipe r in R:
    C_tag = mapped aspect-direction tags of r
    if |C_tag| == 0 or |C_tag| > B:  discard
    A = recipes satisfying all tag constraints in C_tag
    C_ing = []
    budget = B - |C_tag|
    if 1 <= |A| <= U:
        Q.append((C_tag, C_ing, A))
        continue
    while |A| > U and |C_ing| < budget:
        best = None
        for each ingredient g appearing in A:
            A_include = recipes in A including g
            A_exclude = recipes in A excluding g
            for each branch A_next in {A_include, A_exclude}:
                score = tuple(max(0, |A_next| - U), -|A_next|)
                if best is None or score < best.score:
                    best = (g, A_next, score)
        if best is None:
            break
        C_ing.append(best.g)
        A = best.A_next
    if 1 <= |A| <= U:
        Q.append((C_tag, C_ing, A))
return Q
\end{lstlisting}

% -----------------------------------------------------------

\section{Question Quality Evaluation Questionnaire}
\label{sec:questionnaire}

\Needspace{10\baselineskip}
\begin{lstlisting}
Imagine that you are using a conversational recipe recommendation system and would like to describe your dietary needs in natural language to receive suitable recipe recommendations. Below are several examples of user questions. Please judge them based on your intuitive impression:

1) Whether these questions sound like something a real user would naturally ask;
2) Whether the needs or constraints expressed in the questions are reasonably combined, for example, whether they fit everyday dietary scenarios, contain obvious conflicts, or appear overly rigid or unnatural.

Naturalness score:
5 = Sounds fully natural and plausible for a real user to ask
4 = Mostly natural, with only minor awkwardness
3 = Understandable but noticeably templated, wordy, or mechanical
2 = Substantially unnatural or unlikely to be phrased this way
1 = Highly unnatural, incoherent, or not a plausible user request

Constraint reasonable score:
5 = Constraints form a coherent and realistic everyday dietary scenario
4 = Coherent but somewhat specific or demanding
3 = Possible, but unusually rigid, arbitrary, or only marginally realistic
2 = Strongly overconstrained, oddly combined, or contains a likely conflict
1 = Clearly contradictory, nonsensical, or implausible as a dietary request
\end{lstlisting}

% --------------------------------------------------------------

\section{Case Study: QA Examples}
\label{sec:qa-examp}

Here, we present examples of questions from RecipeADQA, KGQA-style baseline, and PFoodReq. For each dataset, we select the highest-scoring, lowest-scoring, and mean-scoring examples from the LLM-based Question Quality Evaluation and show their scores in Naturalness and Rationality.

\subsection{KGQA-style}
\begin{enumerate}
\item \textbf{Question:} Find low-sodium dishes containing prepared horseradish but not black grapes, and make sure they have calories per serving at most 1236.68 kcal and have total fat per serving at most 45.04 g.\\
\textbf{Example Answers:} Sesame-Crusted Steaks With Buttermilk-Horseradish Sauce; Veal Scaloppine (Or Chicken) With Tomato Cream Sauce; Roasted New Potatoes With Lemon Horseradish; \emph{and 91 more}\\
\textbf{LLM score:} 4,5

\item \textbf{Question:} Recommend low-sodium recipes with mint, without meal and prepared pizza crust, that have total carbohydrates per serving at most 138.51 g and have calories per serving between 1327.63 and 434360.2 kcal.\\
\textbf{Example Answers:} Slow-Cooked Harissa Lamb; Lemon-Mint Panna Cotta\\
\textbf{LLM score:} 2,2

\item \textbf{Question:} Show me high-protein recipes containing salt, excluding bourbon whiskey, that have saturated fat per serving at least 31.73 g, have sodium per serving between 0.0 and 5589.27 mg, and have total carbohydrates per serving at least 5.2 g.\\
\textbf{Example Answers:} California Monterey Jack and Blue Cheese Log; Better Than Better Cheddar; Southern Macaroni and Cheese; \emph{and 106 more}\\
\textbf{LLM score:} 1,1 
\end{enumerate}

\subsection{PFoodReq-style}
\begin{enumerate}
\item \textbf{Question:} Could you recommend dinner-party dishes which do not contain Amaretto, and have fresh basil, and do not have tuna, and contain protein with desired range 5.0 g to 30.0 g?\\
\textbf{Example Answers:} Caramelized-Garlic Chicken; Winner's Circle of Stuffed Mushrooms\\
\textbf{LLM score:} 4,4 \\

\item \textbf{Question:} What are low carbohydrates squid dishes which don't consist of cayenne, and have carrots, and do not have coarse grain mustard, and contain calories from carbohydrates with desired range 0.0 \% to 45.0 \%?
\textbf{Example Answers:} Florentine Stewed Calamari With Swiss Chard; Fourth Street Grill's Cioppino; Kai Yat Sai Talay (Thai Omelette With Seafood); \emph{and 4 more}\\
\textbf{LLM score:} 2,3 \\

\item \textbf{Question:} What are birthday recipes that do not consist of ingredient vegan margarine and are high protein, and have dog food, and do not have coconut yogurt, lemon Jell - O gelatin, and contain calories from saturated fat with desired range 5.0 \% to 20.0 \%?\\
\textbf{Example Answers:} Simple Dog Birthday Cake\\
\textbf{LLM score:} 1,1

\end{enumerate}

\subsection{RecipeADQA}
\begin{enumerate}
\item \textbf{Question:} What recipes for dessert, picnic, and summer suitable for children that are no bake can I make?\\
\textbf{Example Answers:} OREO Ice Cream Pie; Blue Frozen Pops\\
\textbf{LLM score:} 5,5 

\item \textbf{Question:} I'm looking for recipes for dessert and holiday that are easy that are low protein and low sodium with a sweet character ready within 4 hours including butter, and rank the results by protein and sodium from lowest to highest.\\
\textbf{Example Answers:} Chocolate Covered Mints; Cranberry-Apple Crisp; Creamy Open-Faced Peach Pie; \emph{and 17 more}\\
\textbf{LLM score:} 4,4

\item \textbf{Question:} What recipes for main course suitable for children that are healthy, low calorie, low cholesterol, low fat, low protein, low saturated fat, and low sodium ready within 4 hours can I make, and rank the results by calories, cholesterol, fat, protein, saturated fat, and sodium from lowest to highest.\\
\textbf{Example Answers:} Roasted Potatoes With Spicy Tomatoes; Congee (Jook): Asian Turkey Rice Soup; Restaurant Plaisir/Sante Ratatouille; \emph{and 11 more}\\
\textbf{LLM score:} 2,2 
\end{enumerate}

% ------------------------------------------------------

\section{Noise-Augmented LLM-Based RAG Recommendation Prompt}
\label{sec:rag}

In the recommendation experiment, the Qwen3-1.7B recommender received a natural-language question and an identifier-blind, randomly ordered candidate pool containing up to 25 correct recipes and sufficient highly similar noise recipes to reach 30 candidates; noise recipes were preferentially selected from recipes that violated exactly one constraint. It returned up to 25 recipes in ranked order. The Prompt~\ref{pmpt:rag} is the exact selection and ranking prompt; \texttt{SELECTION\_CAP} was set to 25.
% 分层实验结果展示了按原始答案数量分层后的Macro F1结果，以排除答案数量分布差异的影响。
The stratification results in Table~\ref{tab:rag_gold} present the Macro F1 scores obtained after stratifying by the number of original (Gold) recipe answers, thereby eliminating the influence of disparities in the distribution of answer counts.

% ----------------------------------------------------
\subsection{Stratification results}
\label{sec:rag-result}

\begin{table}[]
\centering
\caption{Stratification Results in F1.}
\label{tab:rag_gold}
\begin{tabular*}{\linewidth}
  {@{\extracolsep{\fill}}lcccc}
    \toprule
    \textbf{Dataset} & \textbf{Gold=1} & 
    \textbf{Gold=2-5} & \textbf{Gold=6-10} & 
    \textbf{Gold=11-25} \\
    \midrule
    PFoodReq-FoodKG
    & 94.36 & 65.08 & 55.23 & 77.94 \\
    PFoodReq-RecipeADKG
    & 90.79 & 80.56 & 81.83 & 87.65 \\
    KGQA-FoodKG
    & 98.86 & 97.75 & 95.93 & 98.79 \\
    KGQA-RecipeADKG
    & 94.07 & 79.39 & 84.12 & 90.34 \\
    \midrule
    RecipeADQA
    & - & - & - & \textbf{99.04} \\
    \bottomrule
\end{tabular*}
\end{table}

\subsection{Prompt}
\label{sec:rag-prom}

\Needspace{19\baselineskip}
\begin{promptfloat}[htbp]
\begin{lstlisting}
[SYSTEM]
You are a recipe recommendation system that ranks candidate recipes for a user question. Return only valid JSON. Do not explain. Do not include reasons. Do not include markdown or any text outside JSON.

[USER]
Select up to <SELECTION_CAP> best recipes from the provided candidates for the user question.

Rules:
- Only choose candidate_code values from this allowed list: <ALLOWED_CANDIDATE_CODES>
- Return at most <SELECTION_CAP> candidate_code values.
- Return fewer than <SELECTION_CAP> when fewer candidates satisfy the question.
- Preserve a best-to-worst order.
- Output only this JSON shape: {"selected_candidate_codes":[...]}
- Do not output reasons or explanations.

User question:
<NATURAL_LANGUAGE_RECIPE_REQUEST>

Candidates:
<ID_BLIND_CANDIDATE_CONTEXTS>
\end{lstlisting}
\caption{Noise-augmented candidate selection and ranking prompt.}
\label{pmpt:rag}
\end{promptfloat}